**Title of the manuscript:** German CheXpert Chest X-Ray Radiology Report Labeler

**Authors:**
- Alessandro Wollek, M.Sc.[1,2],
- Sardi Hyska, MD. [3]
- Thomas Sedlmeyr, B.Sc.[*,1,2]
- Philip Haitzer, B.Sc. [*,1,2]
- Johannes Rueckel, MD. [3,4]
- Bastian O. Sabel, MD. [3]
- Michael Ingrisch, PhD. [3]
- Tobias Lasser, PhD. [1,2]

[*] Thomas Sedlmeyr and Philip Haitzer contributed equally to this work.

[1] Munich Institute of Biomedical Engineering, Technical University of Munich, Garching b. München, Bavaria, Germany

[2] School of Computation, Information and Technology, Technical University of Munich, Garching b. München, Germany

[3] Department of Radiology, University Hospital, LMU Munich, Munich, Germany

[4] Institute of Neuroradiology, University Hospital, LMU Munich, Munich, Germany

**Work originated from:** Munich Institute of Biomedical Engineering, Technical University of Munich, Boltzmannstr. 11, Garching b. München, 85748, Bavaria, Germany.

**Corresponding Author:**

Alessandro Wollek
- Phone: +49 89 289 10840
- Email: alessandro.wollek@tum.de
- Address: Boltzmannstr. 11, Garching b. München, 85748, Bavaria, Germany

**Funding:**

The research for this article received funding from the German federal ministry of health's program for digital innovations for the improvement of patient-centered care in healthcare [grant agreement no. 2520DAT920].

# Abstract

**Purpose:**

The aim of this study was to develop an algorithm to automatically extract annotations from German thoracic radiology reports to train deep learning-based chest X-ray classification models.

**Materials and Methods:**

An automatic label extraction model for German thoracic radiology reports was designed based on the CheXpert architecture. The algorithm can extract labels for twelve common chest pathologies, the presence of support devices, and "no finding". For iterative improvements and to generate a ground truth, a web-based multi-reader annotation interface was created. With the proposed annotation interface a radiologist annotated 1086 retrospectively collected radiology reports from 2020 – 2021 (data set 1). The effect of automatically extracted labels on chest radiograph classification performance was evaluated on an additional, in-house pneumothorax data set (data set 2), containing 6434 chest radiographs with according reports, by comparing a DenseNet-121 model trained on extracted labels from the associated reports, image-based pneumothorax labels, and publicly available data, respectively.

**Results:**

Comparing automated to manual labeling on data set 1: "mention extraction" class-wise F1 scores ranged from 0.8 to 0.995, the "negation detection" F1 scores from 0.624 to 0.981, and F1 scores for "uncertainty detection" from 0.353 to 0.725. Extracted pneumothorax labels on data set 2 had a sensitivity of 0.997 [95 % CI: 0.994, 0.999] and specificity of 0.991 [95 % CI: 0.988, 0.994]. The model trained on publicly available data achieved an area under the receiver operating curve (AUC) for pneumothorax classification of 0.728 [95 % CI: 0.694, 0.760], the model trained on automatically extracted labels 0.858 [95 % CI: 0.832, 0.882] and on manual annotations 0.934 [95 % CI: 0.918, 0.949], respectively.

**Conclusion:**

Automatic label extraction from German thoracic radiology reports is a promising substitute for manual labeling. By reducing the time required for data annotation, larger training data sets can be created, resulting in improved overall modeling performance. Our results demonstrated that a pneumothorax classifier trained on automatically extracted labels strongly outperformed the model trained on publicly available data, without the need for additional annotation time and performed competitively to manually labeled data.

**Key points:**

- **Developed algorithm for automatic German thoracic radiology report annotation.**
- **Automatic label extraction is a promising substitute for manual labeling.**

- **Classifier trained on extracted labels outperformed model trained on publicly available data.**

# Introduction

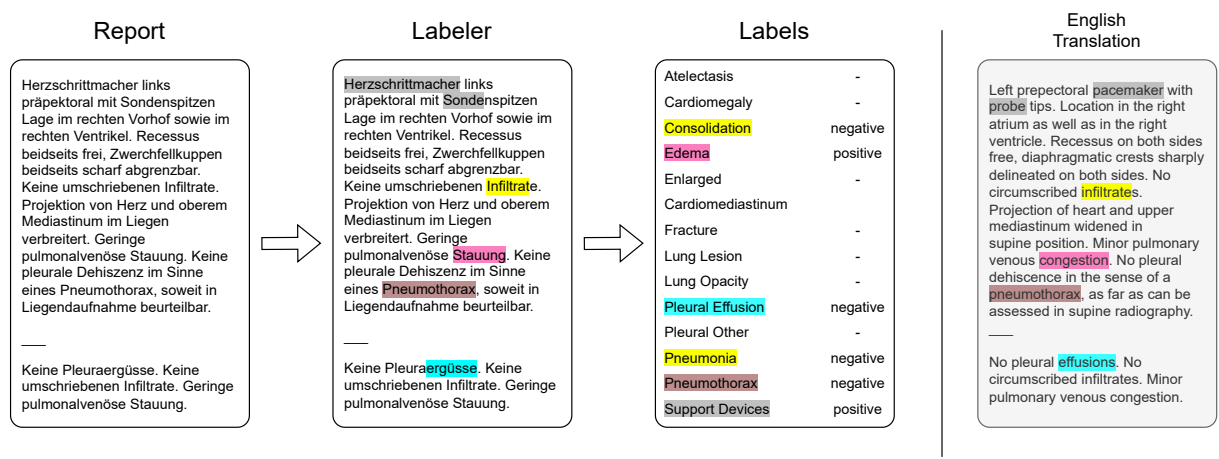

**Figure 1: Automated labeling of German thoracic radiology reports. A report is passed to the report labeler and converted to 14 labels, motivated by CheXpert. The labeler detects each class according to class-specific phrases and converts them to positive, negative, or uncertain labels.**

Chest X-rays are a frequently used and essential tool for detecting lung pathologies, like pneumothorax [1,2]. The accurate interpretation of chest X-rays can be essential for the early detection, timely diagnosis, and effective treatment of these conditions. However, due to the large number of radiological images, radiology departments in many countries and regions are understaffed or overworked, ultimately risking the quality of care [3–5].

Recently, deep learning models used in decision support systems have achieved performance levels in chest X-ray diagnosis of pathologies like pneumonia that are comparable to those of radiologists [6,7]. The integration of such models into clinical systems could reduce repetitive work, decrease workload, and improve the diagnostic accuracy of radiologists.

One of the reasons for the recent surge of innovation based on deep learning models is the availability of large data sets. For example, the release of the ImageNet data set and the according image classification competition, led to huge improvements in the computer vision domain [8–11]. Similarly, the release of the chest X-ray 14 data set [12] sparked the development of chest X-ray classification models like CheXnet [13]. While the number of images used by modern deep learning architectures increased over the years, large publicly available data sets required for new architectures such as Vision Transformers [14] are missing in radiology, limiting the use of advanced models [15] and inhibiting advances in automated chest X-ray diagnosis.

Radiology departments around the world create large amounts of chest X-ray image data with corresponding reports during clinical routine. Despite the existence of huge numbers of radiological imaging studies and their radiological reports stored in the Picture Archiving and Communication Systems (PACS) of numerous clinics, only few are used for the development of new deep learning models, due to missing infrastructure, data privacy considerations, and required time, among others.

Unlike commonly used image data sets, such as ImageNet, chest X-ray data sets obtained from clinical routine require expert annotation due to the specialized knowledge required to understand the images. This annotation task falls on radiologists, who possess the necessary training and expertise to accurately interpret the X-rays. While decision support systems for chest X-ray diagnosis aim to reduce the workload of radiologists, a significant challenge arises from the need for radiologists to perform the time-consuming task of data annotation. This creates a "chicken-and-egg" problem, where the development of decision support systems depends on large, annotated data sets, yet creating these data sets requires significant time and effort from radiologists.

To reduce the amount of time needed for data annotation, natural language processing systems have been created for extracting structured labels from free-text radiology reports. Such systems can be primarily categorized as rule-based or deep learning-based approaches, each of which has its own benefits and limitations. Rule-based systems, for instance, are easier to implement, require no computationally intensive training, provide higher explainability, and can be easily updated with new rules and classes by anyone. On the other hand, deep learning-based approaches primarily rely on large language models, and thus have the potential to produce more accurate label predictions but require more computational resources and larger (manually) annotated data sets. Furthermore, they can only be developed and maintained by experts.

Recent public chest X-ray data sets such as chest X-ray 14, CheXpert [16] and MIMIC-CXR [17] were created by converting existing radiological reports to class labels automatically using rule-based systems. For example, the CheXpert labeler converts an existing report to the thirteen classes: atelectasis, cardiomegaly, consolidation, edema, enlarged cardiomediastinum, fracture, lung lesion, lung opacity, pleural effusion, pleural other, pneumonia, pneumothorax, support devices, and an additional "no finding" class. To minimize development time, the CheXpert labeler was used to annotate the MIMIC-CXR data set as well. Moreover, this labeler has been adapted and ported to process reports in other languages, such as Brazilian [15] and Vietnamese [18]. The process of labeling consists of three stages: In the first stage, mention extraction, the labeler scans the input text for phrases defined in class-specific lists. For example, the phrase list for pneumothorax contains phrases such as "pneumothorax" and "pleural dehiscence". Next, extracted mentions are classified as positive, negative, or uncertain during the second stage (mention classification). Finally, an observation label is created by aggregating all its mentions together (mention aggregation). If a report happens to mention no observation, except support devices, the report is instead labeled as "no finding".

For German radiology reports, Nowak et al. investigated different approaches for training a deep-learning based labeling model [19]. In contrast to the CheXpert labeler, their model predicted only the six observations: pulmonary infiltrates, pleural effusion, pulmonary congestion, pneumothorax, regular position of the central venous catheter (CVC) and misplaced position of the CVC. So far, neither source code nor model weights were released.

In this work, we propose an automatic labeler for German thoracic reports based on the CheXpert algorithm (shown in Figure 1). Our contributions are:

- We created a rule-based labeling algorithm for converting German thoracic radiology reports to CheXpert labels.
- We propose a web-based annotation tool for radiologists to adapt the labeler to new phrases used in a specific clinic and create a ground truth data set.
- We demonstrated that our proposed labeler performs similarly to radiological report labelers in other languages. In addition, we showed that a pneumothorax classifier trained on weakly labeled data outperforms models trained solely on publicly available data, and competitively to manually labeled data.

Our code is publicly available at https://gitlab.lrz.de/IP/german-chexpert.

# Materials and Methods

**Data Collection**

We retrospectively identified thoracic radiology reports from 2020 to 2021 in our institutional PACS and randomly selected 900 reports for the creation of a reference standard and 186 reports for phrase collection and development. In the following, we refer to this data set as data set 1 (DS 1). Initially, two radiologists, one board-certified radiologist with more than ten years' experience (B.S), and one first year radiology resident (S.H.) from Klinikum der Ludwig-Maximilians-Universität München compiled a list of common phrases for each of the fourteen CheXpert classes. During the following data annotation process the list of phrases was expanded, including positive, negative, and uncertain phrases.

**Data Annotation**

| Dataset | Data Set 1 (DS 1) | | | | | | Data Set 2 (DS 2) | | | | | |
| --- | --- | --- | --- | --- | --- | --- | --- | --- | --- | --- | --- | --- |
| Split | Development | | | Test | | | Training | | Validation | | Test | |
| Reports | 186 | | | 900 | | | 4507 | | 660 | | 1267 | |
| Class | P | U | N | P | U | N | P | N | P | N | P | N |
| Atelectasis | 29 | 17 | 1 | 203 | 50 | 2 | - | - | - | - | - | - |
| Cardiomegaly | 34 | 56 | 41 | 166 | 338 | 248 | - | - | - | - | - | - |
| Consolidation | 17 | 28 | 115 | 210 | 23 | 552 | - | - | - | - | - | - |
| Edema | 61 | 3 | 74 | 259 | 11 | 478 | - | - | - | - | - | - |
| Enlarged Cardiom. | 39 | 42 | 52 | 206 | 273 | 277 | - | - | - | - | - | - |
| Fracture | 11 | 1 | 12 | 61 | 4 | 75 | - | - | - | - | - | - |
| Lung Lesion | 11 | 1 | 1 | 37 | 11 | 12 | - | - | - | - | - | - |
| Lung Opacity | 31 | 27 | 112 | 275 | 20 | 484 | - | - | - | - | - | - |
| No Finding | 24 | - | - | 121 | - | - | - | - | - | - | - | - |
| Pleural Effusion | 72 | 7 | 90 | 411 | 49 | 390 | - | - | - | - | - | - |
| Pleural Other | 11 | 3 | - | 53 | 18 | 1 | - | - | - | - | - | - |
| Pneumonia | 4 | 48 | 114 | 52 | 142 | 578 | - | - | - | - | - | - |
| Pneumothorax | 27 | 1 | 147 | 62 | 11 | 786 | 1122 | 3385 | 204 | 456 | 326 | 941 |
| Support Devices | 108 | - | 17 | 523 | 2 | 101 | - | - | - | - | - | - |

**Table 1:** Data sets with data splits and annotated classes used in this study. Data set 1 class annotations were acquired using our proposed annotation interface from free text reports. Data set 2 class annotations were acquired from reports and radiographs [20]. Enlarged Cardiom. = Enlarged Cardiomediastinum, P = Positive, U = Uncertain, N = Negative.

To make the labeling of data set 1 as efficient and accurate as possible, we built a multi-user web-based labeling interface. The design and implementation respect patient data privacy by running the process locally in a secure environment.

The annotation tool, shown in Figure 2, displays the view position and report text on the left side of the screen, with four selectable labeling options available per pathology on the right. These options conform to the original CheXpert architecture and include positive, negative, uncertain, and none, which is used if the specific class was not mentioned. Radiologists can add new class-specific phrases by selecting "add new" and mark and comment on a report for later review. Before saving the annotations, the application highlights the phrases that were recognized by the labeler but were marked as "none" and prompts for a phrase if a class was selected during annotation, but not recognized by the labeler, thereby improving the phrase lists.

To evaluate the labelers performance and expand the class pattern list, one first year radiology resident (S.H.) from Klinikum der Ludwig-Maximilians-Universität München annotated the 1086 randomly selected radiology reports of data set 1 using our proposed annotation interface. The resulting class distribution is listed in **Table 1**.

Figure 2: Report annotation web interface. Top: On the left side view position and report are displayed, on the right the 14 labels can be selected. Additionally, new phrases can be added by clicking "ADD NEW" and a report can be marked for later inspection. Bottom: After clicking "SAVE" the tool highlights the matching phrases with their corresponding labels and asks for a phrase when the selected class was not found by the labeler. Clicking "SAVE" again will save the annotation and load the next report.

**Report Labeler**

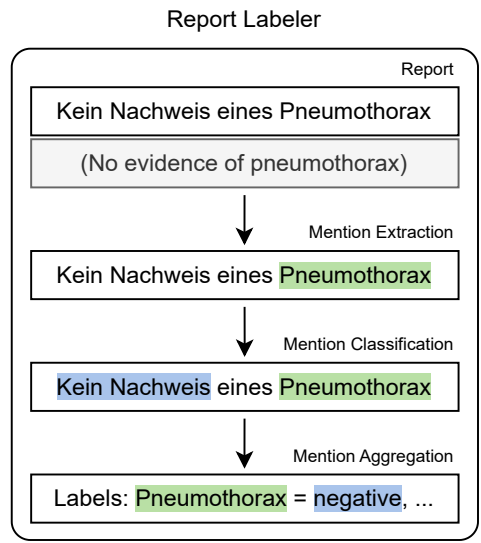

**Figure 3: Labeling flow from our proposed report labeler based on the CheXpert architecture. The report is first matched against a set of class-specific phrases. Afterwards, each match is classified as positive, negative, or uncertain. If the report did not match any phrase it is labeled as no finding in the final stage. English translation provided below the German report excerpt.**

In German radiology reports, two distinct types of negations were identified: expressions that contain phrases like "nicht" or "kein" ("no", "not") and are observation-independent, which can be resolved by the German NegEx algorithm [21]. The other class comprises medical terms that lack any negations but convey the lack of an observation, for example, "Herz normal groß" ("regular heart size"). As the CheXpert architecture addresses only negated observations, we extended the architecture by using multiple phrase files (positive, negative, uncertain) per observation.

As the original mention classification stage depends on an extensive rule set created for English report texts, our labeler utilizes a modified version of the German NegEx algorithm to classify German mentions instead. In the first step, the labeling algorithm identifies negation phrases such as "kann ausgeschlossen werden" ("can be excluded"), and uncertainty phrases, such as "unwahrscheinlich" ("unlikely"), based on a set of rules and marks them as pre- or post-negation/uncertainty phrases.

To identify whether the classification of a mention is affected by negation/uncertainty terms, a cut-off radius determines how many words before and after the mention are taken into consideration, following the German NegEx algorithm. If the relevant region around the mention does not contain any known negation/uncertainty phrases, the mention is classified as positive. If either a pre-negation or post-negation is found near the mention, it is classified as negative. Finally, if there is an uncertainty phrase in the surrounding region, the mention is classified as uncertain.

To form the final label for each observation, the results from mention classification are aggregated as shown in Figure 4. The following rules are applied to derive the labels:

1. Observations with at least one positive mention are assigned a positive label.
2. Observations with no positive mentions and at least one uncertain mention, are labeled as uncertain.
3. Observations with no positive or uncertain mention or at least one negative mention are classified as negative.

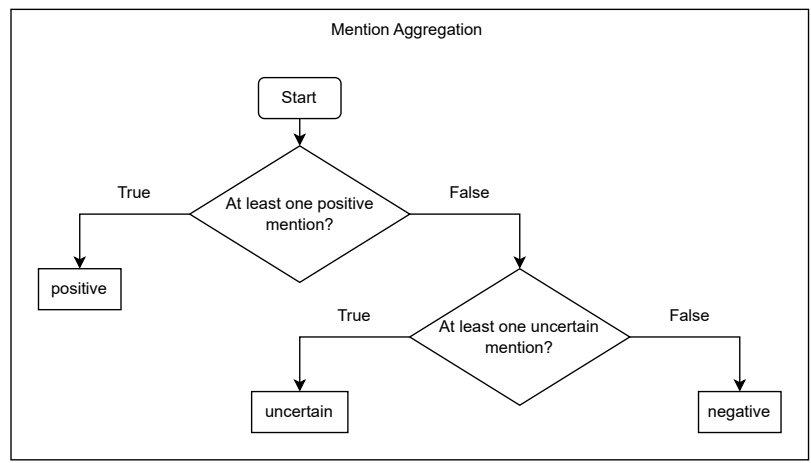

**Figure 4: Derivation of class labels by aggregating all classified mentions per observation. Since an observation can be mentioned multiple times in a report, they must be aggregated for classification.**

The "no finding" label follows a different logic. Initially, a report is labeled as "no finding". The label is changed to negative if any of the other observations (excluding "support devices") are labeled as positive or uncertain.

The main benefit of automated label extraction is time savings. Our proposed algorithm features low memory consumption and enables parallel labeling of multiple reports using multi-threading. Using twelve threads the algorithm labeled 100 reports on average in 1.84 s ± 27.3 ms on a workstation equipped with an Intel i7-6800K CPU with a clock speed of 3.40GHz.

**Label Extraction (DS 1)**

Label extraction performance was measured by comparing extracted and annotated labels on DS 1 on three tasks: mention extraction, negation detection, and uncertainty detection. Regarding the mention extraction task, unlabeled findings ("none") were considered as negative, annotated ("positive", "negative", or "uncertain") as positive. For negation detection, findings annotated as negative were considered as positive, others as negative. For uncertainty detection, annotations were classified analogously. The phrase lists were optimized on the development subset of DS 1. Phrases that were collected during the test subset annotation were discarded to avoid overfitting.

**Pneumothorax Classification (DS 2)**

To measure the effect of automatically extracted labels on downstream model training and classification performance we extracted pneumothorax labels from the radiology reports of an additional internal data set [20]. In the following, we refer to this data set as data set 2 (DS 2). This data set consists of 6434 frontal chest radiographs and their reports, out of which 1568 have been labeled as pneumothorax. Unlike DS 1, the labels are based on radiographs rather than solely reports, and as such, no uncertainty or "none" label are available. We converted the extracted labels to binary labels by considering uncertain cases as positive. For comparison, we applied the same conversion to DS 1 labels and annotations. Additionally, "none" annotations were considered as negative.

We used a DenseNet-121 pre-trained on ImageNet as backbone for our network. We replaced the final fully connected layer with a single output when fine-tuned on DS 2. We replaced the final softmax activation with a sigmoid. We used ADAM with a learning rate of 0.003 and a batch size of 32 and trained for 10 epochs. For our experiments, we selected the best checkpoint based on the validation area under the receiver operating characteristic curve (AUC). All images were normalized according to the ImageNet mean and standard deviation and resized to 224x224 pixels. For data augmentation we applied ten-crop. For our experiments, we compared a DenseNet-121 fine-tuned on the chest X-ray 14 data set (CheXnet) and fine-tuned on DS 2. When fine-tuning on DS 2 we trained with either radiologists' annotations (annotated) or automatically extracted labels (extracted).

**Statistical Evaluation**

We evaluated the labeler's performance using F1 score, precision, and recall regarding mention extraction, negation detection, and uncertainty detection by comparing the extracted to the annotated labels from DS 1. We evaluated pneumothorax classification performance using receiver operating characteristics (ROC) and AUC. Because our study is exploratory and involves multiple comparisons, we refrained from providing P values and provide 95 % confidence intervals calculated using the non-parametric bootstrap method with 10,000-fold resampling at the image level. The labeler performance with respect to the binary pneumothorax labels of DS 2 was measured using sensitivity and specificity. For comparison of DS 1 and DS 2, we converted DS 1 labels and annotations to binary labels and measured sensitivity and specificity. The statistical analyses in this study were done using NumPy version 1.24.2 and Scikit-Learn version 1.2.2.

# Results

**Label Extraction (DS 1)**

| Data Set 1 Findings | Mention Extraction | | | Negation | | | Uncertainty | | |
|---|---|---|---|---|---|---|---|---|---|
| | F1 | R | P | F1 | R | P | F1 | R | P |
| Atelectasis | 0.968 | 0.96 | 0.976 | N/A | N/A | N/A | 0.648 | 0.7 | 0.603 |
| Cardiomegaly | 0.813 | 0.71 | 0.952 | 0.627 | 0.528 | 0.771 | 0.683 | 0.551 | 0.898 |
| Consolidation | 0.933 | 0.919 | 0.947 | 0.884 | 0.802 | 0.984 | 0.4 | 0.609 | 0.298 |
| Edema | 0.993 | 0.996 | 0.991 | 0.965 | 0.941 | 0.989 | 0.48 | 0.545 | 0.429 |
| Enlarged Cardio-mediastinum | 0.867 | 0.807 | 0.937 | 0.678 | 0.569 | 0.84 | 0.725 | 0.607 | 0.902 |
| Fracture | 0.838 | 0.856 | 0.821 | 0.713 | 0.554 | 1.0 | N/A | N/A | N/A |
| Lung Lesion | 0.8 | 0.833 | 0.769 | 0.917 | 0.917 | 0.917 | 0.385 | 0.455 | 0.333 |
| Lung Opacity | 0.92 | 0.915 | 0.926 | 0.851 | 0.743 | 0.994 | 0.364 | 0.6 | 0.261 |
| No Finding | 0.238 | 1.0 | 0.135 | N/A | N/A | N/A | N/A | N/A | N/A |
| Pleural Effusion | 0.99 | 0.985 | 0.995 | 0.948 | 0.938 | 0.958 | 0.5 | 0.429 | 0.6 |
| Pleural Other | 0.864 | 0.792 | 0.95 | N/A | N/A | N/A | 0.8 | 0.778 | 0.824 |
| Pneumonia | 0.902 | 0.829 | 0.988 | 0.862 | 0.771 | 0.976 | 0.705 | 0.612 | 0.833 |
| Pneumothorax | 0.995 | 0.999 | 0.991 | 0.981 | 0.978 | 0.985 | 0.353 | 0.273 | 0.5 |
| Support Devices | 0.939 | 0.92 | 0.96 | 0.842 | 0.762 | 0.939 | N/A | N/A | N/A |

**Table 2: F1 Score, precision and recall for the three evaluation tasks of our report labeler: mention extraction, negation detection, and uncertainty detection for each finding. Labels were extracted from DS 1 and compared to manual annotations. F1 = F1 Score, R = Recall, P = Precision.**

The mention extraction, negation detection, and uncertainty detection results are shown in **Table 2**. Excluding the special case "no finding", mention extraction F1 score ranged from 0.8 to 0.995, negation detection F1 score from 0.624 to 0.981, and the uncertainty detection F1 score from 0.353 to 0.725. The special case "no finding" covers both reports that describe a normal chest radiograph and is the default label when the labeler does not find anything. Since blank "none" labels are considered negative for the mention extraction task, the precision reflects the labeler not finding any mention in the report. Results marked as "N/A" have insufficient samples for calculation.

Commonly, chest X-ray classification models are trained on binary labels. Following Irvin et al. [16], we treat uncertain labels as positive and obtain sensitivity and specificity results as reported in **Table 3**.

|  | Data Set 1 | |
|---|---|---|
| **Findings** | **Sensitivity** | **Specificity** |
| Atelectasis | 0.944 [0.915-0.970] | 0.988 [0.978-0.995] |
| Cardiomegaly | 0.680 [0.639-0.721] | 0.909 [0.880-0.936] |
| Consolidation | 0.952 [0.923-0.978] | 0.892 [0.868-0.914] |
| Edema | 0.970 [0.948-0.989] | 0.946 [0.928-0.963] |
| Enlarged Cardiomediastinum | 0.767 [0.727-0.803] | 0.793 [0.754-0.831] |
| Fracture | 0.954 [0.897-1.000] | 0.959 [0.945-0.972] |
| Lung Lesion | 0.792 [0.667-0.900] | 0.986 [0.978-0.993] |
| Lung Opacity | 0.979 [0.962-0.993] | 0.859 [0.831-0.886] |
| No Finding | 0.736 [0.653-0.813] | 0.983 [0.974-0.991] |
| Pleural Effusion | 0.965 [0.947-0.981] | 0.968 [0.951-0.984] |
| Pleural Other | 0.789 [0.688-0.881] | 0.998 [0.994-1.000] |
| Pneumonia | 0.874 [0.825-0.920] | 0.977 [0.966-0.987] |
| Pneumothorax | 0.819 [0.727-0.904] | 0.979 [0.969-0.988] |
| Support Devices | 0.902 [0.876-0.927] | 0.906 [0.876-0.935] |

|  | Data Set 2 | |
|---|---|---|
| **Findings** | **Sensitivity** | **Specificity** |
| Pneumothorax | 0.997 [0.994, 0.999] | 0.991 [0.988, 0.994] |

**Table 3: Sensitivity and specificity for the extracted labels compared to the reference annotations on DS 1 and DS 2 with corresponding 95 % confidence intervals. To create binary labels, uncertain labels/annotations were considered positive, "none" negative.**

**Pneumothorax Label Extraction (DS 2)**

The labeler extracted pneumothorax labels from DS 2 reports with a sensitivity of 0.997 [95 % CI: 0.994, 0.999] and specificity of 0.991 [95 % CI: 0.988, 0.994], see **Table 3**. Differences between pneumothorax sensitivity and specificity on DS 1 and DS 2 can be explained by the underlying annotation. Uncertain DS 1 annotations were considered as positive, missing ("none") annotations as negative.

**Pneumothorax Classifier**

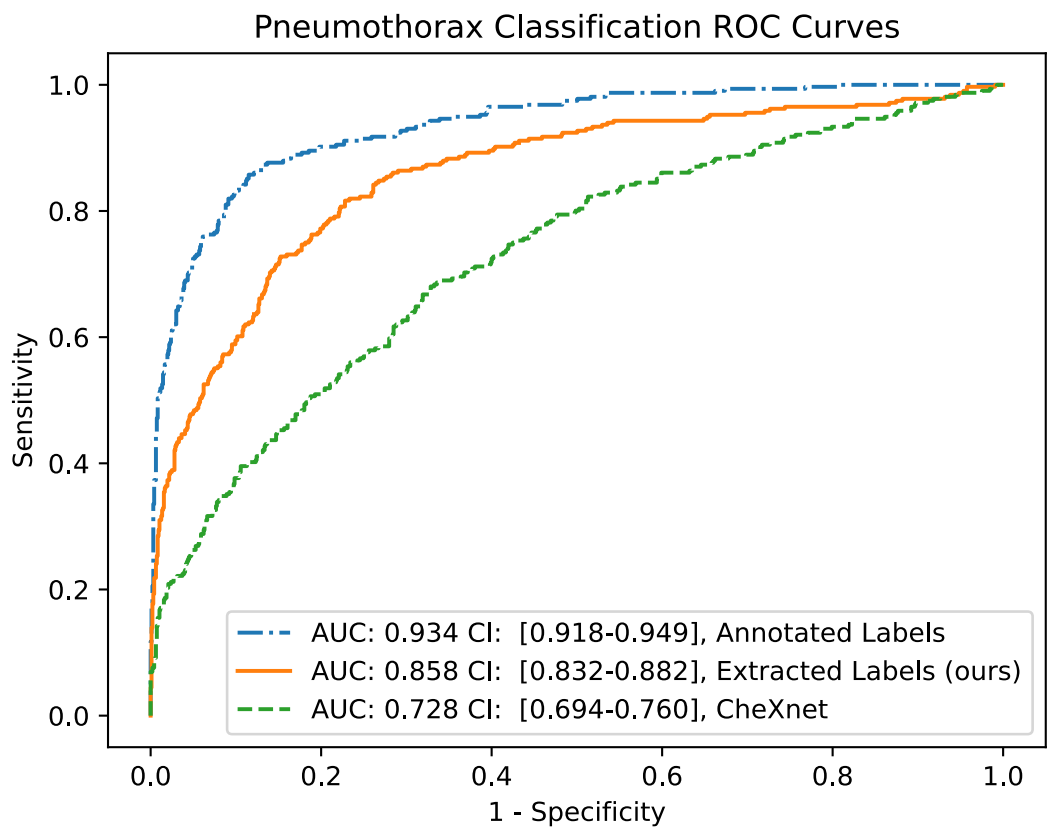

**Figure 5: Receiver operating characteristic (ROC) curves and areas under the ROC curve (AUC) for pneumothorax classification on chest radiograph on our internal data set (DS 2). The model was trained on public data (CheXnet), on the DS 2 training data with either manual annotation (Annotated Labels) or labels extracted using our report labeler (Extracted Labels).**

The ROC curves and corresponding AUC values for the pneumothorax classification models trained on our internal data set with manually annotated labels or extracted labels and trained on the chest X-ray 14 data set are shown in Figure 5. Training with manually annotated labels from multiple readers performed best with an AUC of 0.934 [95 % CI: 0.918, 0.949], followed by the model trained with labels extracted automatically with our labeler with an AUC of 0.858 [95 % CI: 0.832, 0.882]. The CheXnet model trained on chest X-ray 14 data performed worst with an AUC of 0.728 [95 % CI: 0.694, 0.760].

# Discussion

In this study, we proposed an automatic label extraction algorithm for German thoracic radiology reports. Our deep learning model trained on extracted labels demonstrated strong improvements compared to the CheXnet model (0.728 vs. 0.858 AUC) and competitive performance compared to training with manually annotated data (0.858 vs. 0.934 AUC), as shown in **Figure** *5*. This indicates a promising alternative to manual annotation of the training data, especially as the training data set size can be easily scaled with our proposed method. We expect better performance with larger training data sets, allowing for the use of more advanced model architectures, as larger training data sets generally improve image classification performance [14].

Although the extracted pneumothorax labels from DS 2 had a high label sensitivity and specificity of over 99 % (see **Table** *3*), the larger classification AUC difference by the deep learning model trained on manual and extracted labels could be explained by the effect of noisier labels, making it harder to generalize. Creating class labels from radiological reports will always be inferior to the additional inspection of the image and a manual annotation. While pneumothorax label specificity is similar on both data sets, the sensitivity is considerably lower on data set 1, with a larger confidence interval. We interpret this difference as the effect of converting uncertain predictions to positives, as the uncertainty detection F1 score is comparatively low (see **Table** *2*). While greater annotation quality resulted in better label extraction performance, it must be balanced with the time required to create such annotations. The results of our work show that our proposed labeler is a promising tool for clinical data scientists to create data sets.

Our label extraction algorithm was successful in identifying corresponding labels in DS 1 across all classes. The results are in line with other methods proposed in the literature [12,16,22]. Based on our experience, collecting labeling phrases using the proposed interface and the labeler results, we assume that the method can be easily applied to radiology reports from other clinics. Hence, additional classes can be incorporated quickly.

Furthermore, the annotation speed can be greatly improved by running the labeler first. Multiple readers could lower the risk of overlooking classes missed by the labeler.

During the process of annotating radiological reports based on the 14 CheXpert class labels, the radiologists commented that not all class labels were equally simple to annotate. In particular, the class "pleural other" was considered too vague for meaningful evaluation. Although the CheXpert labels were chosen based on the glossary of terms for thoracic imaging from the Fleischner Society [23], some of these labels lacked clear definitions, which could lead to inconsistent annotation, particularly when multiple annotators are involved, especially the "uncertain" classification is arguably too vague to be effectively used for modelling. To address these issues, future work could leverage the proposed annotation tool to refine and expand the CheXpert classes, ensuring that the labels are clearly defined and precise.

Images from a single clinic cannot be representative for the global population. Most chest X-ray data sets that are currently publicly available, such as Chest X-ray 14, CheXpert, or MIMIC-CXR stem from U.S. clinics. By establishing a set of shared class labels and developing chest X-ray report labels for other languages, models build on multi-institutional data sets will be more robust and general. We hope that our work motivates further research in other languages.

One limitation of our work is that we evaluated the effect of automatically extracted labels on chest X-ray classification performance only for the pneumothorax case, not for others. Future work will evaluate the model on all fourteen classes. Another limitation is that the proposed labeler cannot handle semantically equivalent words due to its rule-based nature. In a follow-up work we plan to replace it with a more sophisticated language model. Finally, we observed that few radiology reports described several images. Hence, extracted labels might refer not to the chest radiograph but another image.

In conclusion, we showed that extracting CheXpert labels automatically from German chest X-ray radiology reports are a promising substitute for manual annotation. A pneumothorax model trained on these extracted labels demonstrated competitive performance compared to manually annotated data.

# Clinical relevance

The presented automatic label extraction model for German thoracic radiology reports and the annotation interface are promising tools for radiologists. They can efficiently annotate large data sets for training deep learning-based chest X-ray classification models. Clinical decision support by such models can reduce the workload on radiologists, resulting in improved productivity, and more importantly, accurate and timely diagnosis of chest pathologies.

## Conflict of Interest

The authors declare that they have no conflict of interest.